\documentclass[sigconf]{acmart}

\usepackage{booktabs} 

\usepackage{enumitem}

\usepackage{float}

\usepackage[ruled]{algorithm2e} 

\usepackage{amsfonts}  
\usepackage{mathtools}
\DeclarePairedDelimiterX{\infdivx}[2]{(}{)}{%
  #1\;\delimsize\|\;#2%
}

\usepackage{color,soul}

\newcommand{\vect}[1]{\boldsymbol{#1}}

\SetAlFnt{\small}
\SetAlCapFnt{\small}
\SetAlCapNameFnt{\small}
\SetAlCapHSkip{0pt}
\IncMargin{-\parindent}

\begin{document}
\title{The Art of Drafting: A Team-Oriented Hero Recommendation System for Multiplayer Online Battle Arena Games}


\author{Zhengxing Chen}
 \affiliation{
   \institution{Northeastern University}
   \city{Boston}
   \state{MA}
   \country{USA}
 }
 \email{czxttkl@gmail.com}

\author{Truong-Huy D Nguyen}
 \affiliation{%
   \institution{Fordham University}
   \city{Bronx}
   \state{NY}
   \country{USA}
 }
 \email{tnguyen88@fordham.edu}
\makeatletter
\def\@copyrightspace{\relax}
\makeatother
\author{Yuyu Xu}
 \affiliation{
   \institution{Northeastern University}
   \city{Boston}
   \state{MA}
   \country{USA}
 }
 \email{yuyuxu@ccs.neu.edu}

\author{Christopher Amato}
 \affiliation{
   \institution{Northeastern University}
   \city{Boston}
   \state{MA}
   \country{USA}
 }
 \email{c.amato@northeastern.edu}
 
 \author{Seth Cooper}
 \affiliation{
   \institution{Northeastern University}
   \city{Boston}
   \state{MA}
   \country{USA}
 }
 \email{scooper@ccs.neu.edu}
 
 \author{Yizhou Sun}
 \affiliation{
   \institution{University of California, Los Angeles}
   \city{Los Angeles}
   \state{CA}
   \country{USA}
 }
 \email{yzsun@cs.ucla.edu}

\author{Magy Seif El-Nasr}
 \affiliation{
   \institution{Northeastern University}
   \city{Boston}
   \state{MA}
   \country{USA}
 }
 \email{m.seifel-nasr@northeastern.edu}




\begin{abstract}
\textit{Multiplayer Online Battle Arena} (MOBA) games have received increasing popularity recently. In a match of such games, players compete in two teams of five, each controlling an in-game avatar, known as \textit{heroes}, selected from a roster of more than 100. The selection of heroes, also known as \textit{pick} or \textit{draft}, takes place before the match starts and alternates between the two teams until each player has selected one hero. Heroes are designed with different strengths and weaknesses to promote team cooperation in a game. Intuitively, heroes in a strong team should complement each other's strengths and suppress those of opponents. Hero drafting is therefore a challenging problem due to the complex hero-to-hero relationships to consider. In this paper, we propose a novel hero recommendation system that suggests heroes to add to an existing team while maximizing the team's prospect for victory. To that end, we model the drafting between two teams as a \textit{combinatorial game} and use Monte Carlo Tree Search (MCTS) for estimating the values of hero combinations. Our empirical evaluation shows that hero teams drafted by our recommendation algorithm have significantly a higher win rate against teams constructed by other baseline and state-of-the-art strategies.
\end{abstract}

%
%



\maketitle

\section{Introduction}

Multiplayer Online Battle Arena (MOBA) is one of the most popular contemporary e-sports. Games such as \textit{League of Legends} (Riot Games) and \textit{DOTA 2} (Valve Corporation) have attracted millions of players to play and watch~\cite{playerbase2014,27million}. In a classic match of such games, two teams, each composed of five players, combat in a virtual game map, the goal of which is to beat the opposite team by destroying their base. Each player controls an in-game avatar, known as \textit{heroes}, to co-operate with other teammates in attacking opponents' heroes, armies, defensive structures, and ultimately base, while defending their own in-game properties. 

The selection of heroes, also known as \textit{pick} or \textit{draft}, takes place before each match starts and alternates between two teams until each player has selected one hero. The alternating order of drafting is ``1-2-2-2-2-1'', meaning that the first team picks one hero, followed by the second team picking two heroes, then the first team picking two heroes, and so on. The process ends with the second team picking their last hero. We refer to the 10 heroes in a completed draft as a \textit{hero line-up}.

Heroes are often designed with a variety of physical attributes and skills, which together add to a team's overall power. Therefore, players need to draft heroes that can enhance the strengths and compensate for weaknesses of teammates' heroes (i.e., \textit{synergy}), while posing suppressing strengths over those in the opponent team (i.e., \textit{opposition}). For example, in DOTA 2, hero \textit{Clockwerk} has high synergy with \textit{Naix} because \textit{Clockwerk} can transport \textit{Naix} to target enemies directly, making up for \textit{Naix}'s limited mobility in fighting. In another example, hero \textit{Anti-Mage}'s mana burn skill reduces an opponent's mana resource, making him a natural opposition to \textit{Medusa}, the durability of whom is completely reliant on how much mana she has.
 
In games like League of Legends and DOTA 2, there are possibly more than 100 heroes that can be picked by a player at the time of drafting. As estimated by~\cite{hanke2017reco}, the number of possible hero line-ups in DOTA 2 is approximately $1.56 \times 10^{16}$. Due to the complex synergistic and opposing relationships among heroes, as described above, and large numbers of follow-up pick possibilities by other players, selecting a suitable hero that can synergize teammates and counter opponents is a challenging task to human players. 


In terms of the gaming experience, failing to pick heroes that fit teammates' currently selected heroes and counter opponent heroes can lead to negative feelings, reducing the joy of playing an otherwise entertaining game and igniting even toxic behaviors. Specifically, with poorly picked heroes, players tend to experience repeated losses, which brings frustration and leads to disengagement in the game~\cite{schoenau2011player,yee2006motivations,sherry2006video, chen2015analytics}. Furthermore, it may lower players' confidence and gives rise to quitting matches early~\cite{shores2014identification}, blaming~\cite{kou2013regulating}, and  cyberbullying~\cite{kwak2015exploring}, all of which negatively impact the experience of not only one player but potentially the whole player community.

Since MOBA games are team-based, we propose a hero pick recommendation system for suggesting the approximated ``optimal'' hero pick for team victory. To this end, we view the drafting between two teams as a \textit{combinatorial game}, characterized as a two-person zero-sum game with perfect information, discrete and sequential actions and deterministic rewards~\cite{browne2012survey}. Under this problem formulation, the goal is to decide at each step an optimal hero pick that leads to the hero line-up with the largest predicted win rate on its team, assuming that both teams behave optimally in the remaining picks. This problem, known as sequential decision making, can be solved by search algorithms. However, in the early or mid stage of the draft, exhaustive search might be prohibitive due to the large branching factor of more than 100 candidate heroes per pick. To tackle the large search space, we propose to use Monte Carlo Tree Search (MCTS), a heuristic search algorithm that efficiently estimates the \textit{long-term value} of each pick by simulating possible following picks until completion. Each hero-lineup is associated with a reward, defined as a predicted team win rate representing the estimated strength of the hero line-up. MCTS then back-propagates the reward to update the values of simulated picks. As more simulations are executed, the estimated values become more accurate, allowing MCTS to identify the next pick optimal for team victory. 

The specific version of MCTS~\cite{kocsis2006bandit} we use, namely Upper Confidence Bound applied to Trees, or UCT, is an \textit{anytime} algorithm, i.e., it has the theoretical guarantee to converge to the optimal pick given sufficient time and memory, while it can be stopped at any time to return an approximate solution. In contrast, previous works either predict player tendency to pick heroes~\cite{summerville2017reco}, or leverage association rules mined from historical hero line-ups as heuristics~\cite{hanke2017reco}. They are not guaranteed to converge to the optimal hero pick for team victory.



In sum, the contributions of our paper are three-fold.
\begin{enumerate}
\item we provide a formal formulation of the drafting process in MOBA games as a combinatorial game;
\item we detail a new hero recommendation approach to solve the above problem using an MCTS algorithm;
\item we report the results of our empirical simulation experiments, demonstrating the superiority of our algorithm over other baseline and state-of-the-arts strategies.
\end{enumerate}

Our paper is structured as follows. We first present background and related works, then introduce the problem formulation and our algorithm. Next, we detail the procedure of our experiments as well as evaluation results. Finally, we conclude our paper and discuss limitations as well as future directions.

\section{Drafting in MOBA Games}
MOBA games have attracted a variety of research thanks to their complexity and design richness. Some research problems of interest include team formation analysis~\cite{pobie1,pobie2,neidhardt2015team,kim2016proficiency}, skill analysis~\cite{zhengxing2016player,Drachen:skill}, and hero pick recommendation systems~\cite{summerville2017reco,hanke2017reco}. 

A typical MOBA match is played by two teams of five players, each of whom controls one hero selected from a pool of more than 100 candidates before the match begins. Each hero has different abilities, making it suitable for different tactical roles or play styles in the game, e.g. dealing long-distance damage, healing teammates, or spearheading with strong shields. Moreover, there exist sophisticated synergistic and oppositional relationships between heroes, which are learned over time by experienced players. Due to such sophistication, the hero drafting phase also becomes a critical component contributing to match outcomes, even though it happens before the real match starts. Previous research has highlighted that interactions between heroes greatly influence match outcomes~\cite{pobie1,Semenov2016,kim2016proficiency}. There also exist many online discussions recommending hero picks within the player communities~\footnote{\url{http://www.weskimo.com/a-guide-to-drafting.html}} \footnote{\url{https://www.reddit.com/r/learndota2/comments/3f9szo/how_to_counter_pick_heroes/}}.

In the drafting process, two teams alternate to pick heroes in a ``1-2-2-2-2-1'' order, during which heroes already selected are visible to both teams. Heroes can only be selected from a fixed pool and no duplication is allowed in the same match. Note that the drafting rules described here are based on ``Ranked All Pick'', a popular match mode in DOTA 2. For simplicity of illustration, we will first focus on this kind of match and its drafting process. There are other more sophisticated mechanics deployed in the drafting process that may affect match outcomes, such as \textit{banning} (i.e., certain heroes can be prohibited from selection by either team), which we will study in the end of our performance evaluation (Section~\ref{sec:extension}).

\section{Related Works}

Previous works on hero pick recommendation can be categorized into two main approaches, based on (1) historical selection frequency, and (2) previous win rate.

In the first category, researchers proposed to recommend heroes that players tend to select~\cite{summerville2017reco}. Essentially, hero pick recommendation is modelled as a sequence prediction problem, in which the next hero to pick is predicted from the sequence of picks made already in the draft. Since the sequence prediction model is trained from pick sequences in historical matches, the predicted hero is ``what is
likely to be picked, not what is necessarily best''~\cite{summerville2017reco}. Therefore, hero recommendation based on such a method may not be optimal for team victory.

In the second category, heuristics are learned and used to seek for heroes that can improve the team expected win rate. \citeauthor{hanke2017reco} proposed to mine association rules~\cite{agrawal1994fast} from historical hero line-up data and use them as the heuristic to recommend heroes~\cite{hanke2017reco}. Here, association rules are hero subsets that appear frequently together either in the winning team or in opposite teams. Any hero contained in the discovered association rules together with the heroes picked already is suggested to be a good candidate to pick next. However, such method does not consider what heroes \textit{will} be picked in the rest of the drafting process, hence this is essentially a myopic, greedy-based approach.
  
Although there are other works using machine learning models to predict match outcomes~\cite{Yang:identifying,Semenov2016}, they do not focus on how to utilize these models for hero recommendation.





\section{Methodologies}\label{sec:methodologies}
In  this  section  we  will  formally  describe  how  a draft between two teams can be cast as a combinatorial game. We will then proceed to presenting how Monte Carlo Tree Search (MCTS) is applied to compute the optimal hero pick.

\subsection{Problem Formulation}\label{sec:probdef}
As we introduced previously, a draft takes place with players from the two teams picking heroes in an alternating order. We assume all players from the same team share the same goal: build a hero line-up with the highest predicted win rate as a team prior to the real match starts. Therefore, although in real life the five players on a team are different individuals, we regard the whole team as one collective player. As such, the draft can be considered as a game played by two players, each representing a team. Following the terminology in DOTA 2, the two players are called \textit{Radiant} and \textit{Dire}, respectively. Without loss of generality, we assume that the problem is formulated with Radiant being the team to pick heroes first. 

More specifically, the draft can be defined as a combinatorial game~\cite{browne2012survey} with the following elements:

\begin{itemize}[leftmargin=*]
\item the number of players: $n = 2$.
\item game states: $S \subset \mathbb{Z}^{N}$. A game state $s \in S$ is an $N$-dimensional vector encoding which heroes have been picked by both players, where $N$ is the number of total distinct heroes. The components of $s$ can take value as one, negative one, and zero, corresponding to a hero being picked by Radiant, Dire, or neither of them, respectively:
\begin{equation}
s_{i}=
\begin{cases}
  1, & \text{hero } i \text{ picked by Radiant} \\
  -1, & \text{hero } i \text{ picked by Dire} \\
  0, & \text{otherwise}
\end{cases}
\label{eqn:sifeature}
\end{equation}
The number of components equal to one (or negative one) cannot exceed five, because each player will eventually select five heroes. As per the rule of drafting, states are fully observable to both players. 
\item the initial game state $s_0 = \vect{0}^N$. $s_0 \in S$ is a blank draft with no hero picked yet, so it is a zero vector. 
\item the terminal states $S_T \subseteq S$, whereby the draft is complete. $S_T$ include those states with exactly five components being one and five components being negative one, denoting a completed draft.
\item the turn function $\rho: S \rightarrow (\text{Radiant}, \text{Dire})$. It decides which player is about to pick in each state. $\rho$ is based on the pick order ``1-2-2-2-2-1'' between the two players.   
\item the set of actions $A$. Each action represents a hero pick by the current player, i.e., (deterministically) changing a zero component of a non-terminal state $s \in S \setminus S_T$ to one (if the picking player is Radiant) or negative one (if the picking player is Dire). There are a finite number of actions and they are applied sequentially according to $\rho$. 
\item the reward function $R: S \rightarrow \mathbb{R}^2$. $R(s)$ outputs a two-dimension reward, with the first component $R^1(s)$ being the reward assigned to Radiant and the second component $R^2(s)$ being the reward assigned to Dire. Since both teams strive to maximize the predicted win rate of their team based on the completed draft, $R$ is only defined for terminal states $s \in S_T$, whereby $R^1(s)=-R^2(s)=w(s)$ with $w(s)$ denoting the predicted team win rate of Radiant. 
\end{itemize}

Designed with the elements above, the draft is regarded as a combinatorial game characterizing the two-person, zero-sum property, perfect information, deterministic rewards, and discrete and sequential actions. 


\begin{figure*}
\centering
\includegraphics[width=0.8\textwidth]{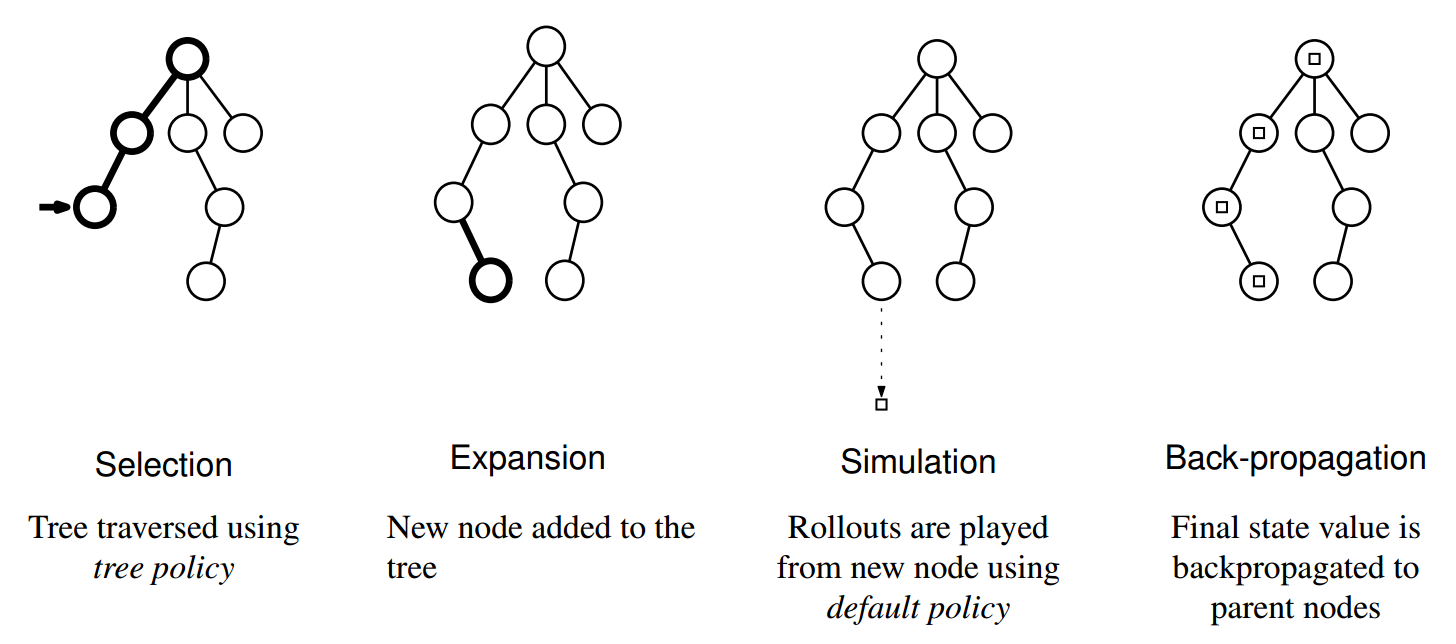}
\caption{Diagram representing the four steps of UCT. Diagram adapted from~\cite{browne2012survey} and~\cite{santos2017monte}}
\label{fig:mctsplot}
\end{figure*}

\subsection{Apply Monte Carlo Tree Search to Find Optimal Pick}\label{sec:mcts}
Many classic games like Go, Chess, and Tic Tac Toe are also combinatorial games. A popular approach to solve combinatorial games is the \textit{minimax} algorithm~\cite{knuth1975analysis}, which finds the optimal action for a player by constructing a complete search tree comprising of all possible actions alternated by each player. Minimax assumes that both parties are playing rationally, i.e., each selecting actions that maximizes their accumulated rewards or minimizing accumulated costs. Since minimax is exhaustive in nature, it does not scale well for games with a large search space such as Go. Therefore, heuristic functions to approximate the actions' values after limited search depth and/or techniques for pruning search trees~\cite{knuth1975analysis} are needed. Even with these techniques, minimax may still be infeasible or not perform well for complex search problems.

As an alternative to minimax, \textit{Monte Carlo Tree Search} (MCTS)~\cite{coulom2006efficient,kocsis2006bandit,nguyen2014bootstrapping} is a class of heuristic search algorithms for identifying near-optimal actions without having to construct the complete search tree. It has attracted much attention in recent years due to successful application to Go-playing programs~\cite{silver2016mastering,silver2017mastering}, General Game Playing~\cite{finnsson2008simulation}, and Real-Time Strategy games~\cite{balla2009uct}. In essence, MCTS involves iteratively building a search tree at each decision state to estimate the values of the state and available actions at that state. The main idea of MCTS in improving the efficiency of search is to prioritize expanding the search trees in the direction of the most promising actions.


A draft in a normal MOBA is regarded as a combinatorial game with a branching factor at least $100$ and depth 10, since popular MOBA games such as League of Legends and DOTA 2 sport more than 100 heroes to be selected for each of the 10 picks in a draft. As the branching factor is very large in this case, which makes minimax hardly a feasible approach, we propose to apply MCTS to compute the optimal pick for the current player.



Specifically, we propose the use of a particular version of MCTS called \textit{Upper Confidence Bound applied to trees} (UCT)~\cite{kocsis2006bandit} for this purpose. The search tree of UCT is built iteratively, with each node representing a state and each directed edge to child nodes representing the action resulting to a next state. It starts with a root node, which, in our case, represents the draft state $s$ right before the current player is to pick the next hero. Then, the search tree is built based on the following four steps per iteration, until time or memory resource allocated is depleted:


\textbf{Selection}. In this step, the algorithm starts from the root node and traverses down the tree to reach a terminal node or an expandable node. A node is \textit{expandable} if it is a non-terminal state and has child nodes unvisited. The tree traversal is done by successively selecting child nodes, following the \textit{tree policy}, which in UCT is based on the \textit{Upper Confidence Bound} (UCB1) criterion~\cite{auer2002finite}:
    \begin{equation}
    \begin{aligned}
    \pi_{UCB1}(s) = \arg\max_a \Big\{ \bar{w} + c \sqrt{\frac{\log n(s)}{n(s, a)}} \Big\},
    \label{eqn:ucb}
    \end{aligned}
    \end{equation}
where $s$ and $a$ refer to a parent node and an action available at that node, respectively; $\bar{w}$ is the average reward received after taking $a$ at $s$; $n(s,a)$ is the number of times $a$ is sampled at $s$, and $n(s)$ is the total number of times $s$ is visited; and $c$ is the exploration term, usually chosen empirically. 

What is implied in Eqn.~\ref{eqn:ucb} is that UCT regards the choice of actions in the selection phase as multi-armed bandit problems: it focuses on \textit{exploiting} the most promising actions to expand next (controlled by $\bar{w}$), while balancing that by \textit{exploring} more neglected branches of the tree (controlled by $c \sqrt{\frac{\log n(s)}{n(s, a)}}$). 
    
\textbf{Expansion}. Unless the reached node from the last step is a terminal state, one of the unvisited actions is randomly sampled and applied to the node. The child node representing the resulting state is added to the tree.

\textbf{Simulation}. From the newly expanded node, the algorithm performs a roll-out until the end of the game. During the roll-out, actions are performed according to a \textit{default policy}, which by default is random sampling. Once the roll-out reaches a terminal state $s \in S_T$, $R(s)$ are calculated.

\textbf{Backpropagation}. The reward is backpropagated from the expanded node to the root node. Statistics on each node, i.e., the average reward and the number of visits, on the path are updated accordingly.

As the number of these four-step iterations increases, the algorithm expands the search tree larger and touches on more states and actions. Intuitively, the larger the search tree is, the better the value approximation of state nodes is. When the algorithm terminates, the action leading to the root's most rewarding child node, i.e., highest $\bar{w}$, is returned. The tree building process of UCT is sketched in Figure~\ref{fig:mctsplot}.

It is proven that when proper parameters are configured and rewards are bounded to the range $[0,1]$, UCT converges to minimax's optimal action at a polynomial rate as the number of iterations grows to infinity~\cite{kocsis2006bandit}. This implies that applying UCT with theoretically sufficient time and memory will eventually converge to the optimal hero for team victory. 

There are two benefits of MCTS in general that makes it suitable for a hero pick recommendation system. First, state value estimation could solely rely on the backpropagation of the reward on terminal states without needing to resort to a hand authored heuristic function. This reduces the amount of domain knowledge required. Second, MCTS is an anytime algorithm, which means tree building can be interrupted when a given time or memory constraint is exceeded and estimated values based on the search tree built so far can be returned immediately. This makes it feasible for our MCTS-based method to be deployed in large-scale online matches under real-world resource constraints.

\subsection{Reward Function as a Win Rate Predictor}\label{sec:rewardfunction}
Now, we describe how the reward, i.e. the team win rate based on a hero line-up, is calculated. 

Following the previous notation, we use $w(s)$ to denote the predicted team win rate from the view of the Radiant player, based on a complete draft $s \in S_T$. This win rate can be computed by a machine learning classification model trained on a large amount of hero line-ups extracted from historical matches. During the training of such a model, the input feature vector is the completed draft as encoded in Eqn.~\ref{eqn:sifeature}; the output is a binary label indicating the observed match outcome (+1 for a win or 0 for a loss, from the view of Radiant). There are two properties making a model desirable for our task: (1) it should return the class probability, i.e., the probability of Radiant drafts winning the game, rather than just a binary prediction; and (2) it should be a non linear-based model that can capture the interrelationships between features (i.e., heroes).

\section{Performance Evaluation}\label{ressection}
In this section, we detail the set up of our simulation-based experiments and demonstrate the effectiveness of UCT by showing that teams following our algorithm to pick heroes can form stronger hero line-ups against teams following other hero pick strategies.

\subsection{Data Collection}
We choose the DOTA2 match dataset collected between February 11, 2016 to March 2, 2016 by Semenov et al.~\cite{Semenov2016}. No major game update affecting the mechanics of the games occurred during the data collection phase. The original dataset contains five million ``Ranked All Pick'' matches, with each match containing hero line-up information and average player skill level (i.e., normal, high, and very high). To further reduce the impact introduced by player skill, 
we extract a subset of matches played by gamers only in the normal skill level. In total, our dataset has 3,056,596 matches, containing 111 distinct heroes. The dataset will be used to train a team win rate predictor as the reward function, as well as provide a basis for our simulations.

\subsection{Win Rate Predictor Training}


To prepare the data for training the team win rate predictor, each match's hero line-up is encoded as a feature vector of length 111 (Eqn.~\ref{eqn:sifeature}) while the match outcome is used as the label. 

We experiment with three classification models, Gradient Boosted Decision Tree (\textbf{GBDT})~\cite{friedman2001greedy}, Neural Network (\textbf{NN})~\cite{bishop2006pattern}, and a generalized linear model, Logistic Regression (\textbf{LR}), for the team win rate predictor. GBDT fits the logit of label probabilities with the outputs of a collection of decision trees learned using boosting~\cite{friedman2000additive}. For the NN model, we use one input layer, one hidden layer, and one output layer. The hidden layer is comprised of a number of rectified linear units (ReLU) which transform the original features non-linearly to feed the output layer. The output layer is a single neuron with the sigmoid activation function $\frac{1}{1+exp(-x)}$. 
LR models the logit of label probabilities as a linear combination of individual features without explicitly modeling the interactions among features. Therefore, although all the three models can predict the label probabilities on new data, GBDT and NN are more sophisticated and able to capture interactions among features. We also test a naive baseline model which always outputs the majority class (\textbf{MC}). In our case, it always predicts a win for Radiant because there are 53.75\%  matches in which Radiant wins.

All hyperparameters like the number of hidden units of NN, the number of trees of GBDT and the regularization penalty of LR are determined through grid search on a 10-fold cross-validation procedure. In each fold, we split the data into the train, validate and test set in an 8:1:1 ratio. Candidate models of the same kind but with different hyperparameters are trained on the train dataset; the best hyperparameters are determined according to the classification accuracy on the validation dataset. The classification performance of the best model will be measured on the test dataset.  

\begin{table}
  \caption{Performance of Team Win Rate Predictors}
  \centering
  \label{tab:auc}
  \begin{tabular}{c@{\hskip 0.5in}c@{\hskip 0.36in}c@{\hskip 0.36in}c}
    \toprule
    Model  & Accuracy & AUC  \\
    \midrule
    MC   & 0.53779        & 0.5       \\
    LR   & 0.63576        & 0.68767    \\
    GBDT & 0.64172        & 0.70142   \\
	NN   & 0.65345        & 0.71437  \\
  \bottomrule
\end{tabular}
\end{table}

We report the accuracy and area under ROC (receiver operating characteristic) curve (AUC) for each kind of model in Table~\ref{tab:auc}, averaged over 10-fold cross validation. NN achieves the best prediction performance in both accuracy and AUC. Additionally, the accuracy and AUC of NN are statistically higher than those of the other models according to a paired, two-tailed Welch's t-test with confidence level $0.05$. Thereby we decide to use NN as the reward function in our simulation experiments later. 

It is worth noting that, the absolute difference between LR and NN is 1.8\% for accuracy and 0.03 for AUC, which may give a wrong impression that match outcomes are accountable by the sum of individual heroes' effects and hero interactions are not as important as we want to emphasize. We find a possible explanation for this phenomenon, that players already tried hard to build closely competitive hero line-ups in the collected matches such that good or bad hero interactions are hard to stand out. If we train the models with additional matches in which players are forced to play random heroes, NN may have a larger edge over LR. We will investigate such an issue in the future.

\subsection{Simulation Setup}
We design four strategies for hero pick recommendation and test their effectiveness in terms of team victory. For every pair of strategies, we conduct 1000 simulations: in each simulation, two teams participate in the drafting process, with each team adopting one of the two strategies. At the end of each draft, we collect the predicted win rate as measure of strength of the built hero line-ups. Finally, we will report the mean of the predicted win rates over the 1000 drafts for each pair of strategies. The procedure of one simulation is summarized in Algorithm~\ref{alg:sim_match}.

\begin{algorithm}
    \SetKwInOut{Input}{Input}
    \SetKwInOut{Output}{Output}
    \SetKwFunction{Armijo}{armijo} 
    \Input{team1, team2, win rate predictor $\mathcal{M}$}
    \Output{$w(s)$}
    \BlankLine
    Initialize a new draft $s:= s_{0}$ 
    \BlankLine
    \While{$s \not\in S_T$} {
    team = $\rho(s)$
    
    \uIf{$s$ equal to $s_0$}{
    
        $a$ = weighted\_sample\_hero()
        
    }
    \Else{
    
        $a$ = team.recommend\_hero($s$)
        
    }

    $s = f(s,a)$
    }
    
    $w(s)$:= predicted by $\mathcal{M}$ with input $s$
    \caption{Simulation of one match}
    \label{alg:sim_match}
\end{algorithm}

\begin{table*}
  \caption{Mean predicted win rate of $UCT_{n, 1}$ (row) vs. $UCT_{n, c}$ (column). Numbers are from the view of the row strategy. Bold cells indicate the best value of $c$ for $UCT_{n,c}$ for each $n$, as they force $UCT_{n,1}$ have the lowest win rate.}
  \label{tab:param_set}
  \centering
  \begin{tabular}{ccccccccc}
    \toprule
      & $UCT_{n,2^{-6}}$ & $UCT_{n, 2^{-5}}$ & $UCT_{n, 2^{-4}}$ & $UCT_{n, 2^{-3}}$ & $UCT_{n, 2^{-2}}$ & $UCT_{n, 2^{-1}}$ & $UCT_{n, 2^0}$ & $UCT_{n, 2^1}$\\
    \midrule
    $UCT_{100, 1}$ & 0.5 & 0.5 & 0.5 & 0.5 & 0.5 & 0.5 & 0.5 & 0.5\\
    $UCT_{200, 1}$ & 0.459 & \textbf{0.435} & 0.442 & 0.444 & 0.447 & 0.464 & 0.5 & 0.540 \\
    $UCT_{400, 1}$ & 0.497 & 0.476 & 0.469 & 0.469 & \textbf{0.457} & 0.469 & 0.5 & 0.534 \\
    $UCT_{800, 1}$ & 0.561 & 0.544 & 0.527 & 0.509 & 0.488 & \textbf{0.485} & 0.5 & 0.525 \\
    $UCT_{1600, 1}$ & 0.606 & 0.577 & 0.563 & 0.545 & 0.510 & \textbf{0.492} & 0.5 & 0.516 \\
  \bottomrule
\end{tabular}
\end{table*}

The four experimented strategies include:
\begin{itemize}[leftmargin=*]
\item \textbf{UCT}: this strategy is what we propose in Section~\ref{sec:methodologies}. We will use $UCT_{n,c}$ to denote a specific UCT version with $n$ iterations allowed and an exploration term $c$. 
\item \textbf{Association Rules (AR)}: this strategy was proposed by Hanke and Chaimowicz~\cite{hanke2017reco}. Two sets of association rules, namely ``ally rules'' and ``enemy rules'', are extracted from the collected matches. Ally rules and enemy rules represent hero subsets that appear frequently in the same winning team or in the opposite team, respectively. At each turn, the strategy looks for the extracted association rules containing both the heroes picked already and the heroes not picked yet. Those not picked yet will be selectively added to a candidate pool, from which the recommended hero will be uniformly sampled. In our implementation, we adopt the best criteria claimed by the authors: association rules are mined with $0.01\%$ minimum support, and the metrics for selectively adding heroes to the candidate pool are ``win rate'' and ``confidence'' for ally rules and enemy rules, respectively. Readers can refer to the original paper~\cite{hanke2017reco} for more details on the approach. 
\item \textbf{Highest Win Rate (HWR)}: each time, pick the hero not selected yet and with the highest win rate, based on frequency counts on our dataset. 
\item \textbf{Random (RD)}: each time, uniformly sample a hero not yet selected. 
\end{itemize}

We do not implement the strategy based on sequence prediction models as proposed by Summerville et al.~\cite{summerville2017reco} because their model requires training on hero pick sequences, which are not readily available in our dataset. In fact, hero pick sequences are not currently downloadable from the official APIs provided by the game and need to resort to third-party organizations. We would implement the strategy in the future should we have access to such data of hero pick sequences.

Some additional implementation details are as follows. For each simulated draft, regardless of the strategy adopted, the first hero is sampled following the probability distribution reflecting how frequently each hero is picked in our dataset. This helps our experiments cover more possible scenarios. To ensure fairness when comparing pairs of strategies, among the 1000 simulations, each strategy is set to start first for half of the simulations. A shared random seed is set for each 500-simulations, to make sure that randomness is the same for both strategies. All strategies follow the rule that no hero can be picked twice. All experiments are conducted on a PC with an Intel i7-3632QM 2.20GHz CPU.

\subsection{Parameter Search for UCT}

We first run simulations to determine the exploration term $c$ for the UCT strategy. We choose $UCT_{n, 1}$ as benchmarks (i.e., UCT with the exploration term ${c=1}$), where ${n=\{100, 200, 400, 800, 1600\}}$. For each $UCT_{n,1}$, we then create multiple $UCT_{n, c}$ strategies to play with, with $c$ ranging from $2^{-6}$ to $2$ at a scaling rate of $2$.

The results are shown in Table~\ref{tab:param_set}. Tuning $c$ is not helpful when UCT is only allowed with 100 iterations. This is because $c$ controls the exploration strength for tree node selection in the \textit{selection} step, which never kicks off within 100 iterations due to the large number of available actions\footnote{We have 111 distinct heroes and 10 turns in a draft, which means applying UCT at any turn will start from a root state with more than 100 possible child nodes (actions) to expand. Since one iteration only expands one new child node, after 100 iterations, the root node is still expandable. Therefore, no selection step will happen. }. We can infer that the best value of $c$ to couple with ${n=\{200, 400, 800, 1600\}}$ is $2^{-5}$, $2^{-2}$, $2^{-1}$ and $2^{-1}$, respectively, because they force $UCT_{n, 1}$ to have the lowest win rate (indicated by the bold cells in Table~\ref{tab:param_set}). Note that there is a general trend that the best $c$ increases as $n$ increases.


\begin{table*}
  \caption{Mean predicted win rate of row strategy  vs. column strategy. Simulations are based on the drafting rules of match mode ``All Ranked''. Numbers are from the view of the row strategy. The strategies are sorted in ascending order by their strengths, from top to bottom, and from left to right. Win rates symmetric to diagonal always sum to one, thus half of them are omitted.}
  \label{tab:mcts}
  \centering
  \begin{tabular}{ccccccccc}
    \toprule
      & $RD$ & $AR$ & $UCT_{100, 1}$ & $UCT_{200, 2^{-5}}$ & $HWR$  & $UCT_{400, 2^{-2}}$ & $UCT_{800, 2^{-1}}$ & $UCT_{1600, 2^{-1}}$ \\
    \midrule
    $RD$ & 0.5 &  &  &  &  &  &  &  \\
    $AR$ & 0.682 & 0.5 &  &  &  &  &  & \\
    $UCT_{100, 1}$ & 0.783 & 0.663 & 0.5 & & &  &  \\
    $UCT_{200, 2^{-5}}$ & 0.897 & 0.833 & 0.712 & 0.5 & & & \\
    $HWR$ & 0.883 & 0.846 & 0.715 & 0.516 & 0.5 \\
    $UCT_{400, 2^{-2}}$ & 0.920 & 0.863 & 0.763 & 0.568 & 0.556 & 0.5   \\
    $UCT_{800, 2^{-1}}$ & 0.928 & 0.878  & 0.776 &  0.591 & 0.593 & 0.524 & 0.5 &  \\
$UCT_{1600, 2^{-1}}$ & 0.930 & 0.880  & 0.787 & 0.606 & 0.611 & 0.539 & 0.513 & 0.5 \\
  \bottomrule
\end{tabular}
\end{table*}

\begin{table*}
  \caption{Average wall time per pick of different strategies (unit: millisecond, match mode: All Ranked).}
  \label{tab:mcts_time}
  \centering
  \begin{tabular}{ccccccccc}
    \toprule
      $RD$ & $AR$ & $UCT_{100, 1}$ & $UCT_{200, 2^{-5}}$ & $HWR$  & $UCT_{400, 2^{-2}}$ & $UCT_{800, 2^{-1}}$ & $UCT_{1600, 2^{-1}}$ \\
    \midrule
     0.02 & 11 & 43 & 96 & 0.1 & 281 & 562 & 1120 \\
    \bottomrule
\end{tabular}
\end{table*}

\begin{table*}
  \caption{Mean predicted win rate of row strategy  vs. column strategy. Simulations are based on the drafting rules of match mode ``Captain Mode''. The table can be viewed in a similar way as Table~\ref{tab:mcts}.}
  \label{tab:mcts_captain_mode}
  \centering
  \begin{tabular}{ccccccccc}
    \toprule
      & $RD$ & $AR$ & $UCT_{100, 2^{-5}}$ & $UCT_{200, 2^{-5}}$ & $HWR$  & $UCT_{400, 2^{-2}}$ & $UCT_{800, 2^{-1}}$ & $UCT_{1600, 2^{-1}}$ \\
    \midrule
    $RD$ & 0.5 &  &  &  &  &  &  &  \\
    $AR$ & 0.693 & 0.5 &  &  &  &  &  & \\
    $UCT_{100, 2^{-5}}$ & 0.817 & 0.686 & 0.5 & & &  &  \\
    $UCT_{200, 2^{-5}}$ & 0.91 & 0.851 & 0.712 & 0.5 & & & \\
    $HWR$ & 0.876 & 0.788 & 0.694  & 0.508 & 0.5 \\
    $UCT_{400, 2^{-2}}$ & 0.929 & 0.886 & 0.770 & 0.574 & 0.570 &  0.5  \\
    $UCT_{800, 2^{-1}}$ & 0.941 & 0.896 & 0.788 & 0.598 & 0.607 & 0.536 & 0.5 &  \\
$UCT_{1600, 2^{-1}}$ & 0.942 & 0.903 & 0.797 & 0.620 & 0.627 & 0.552 & 0.519 &  0.5 \\
  \bottomrule
\end{tabular}
\end{table*}

\subsection{Simulation Results}
Based on the results from the parameter search, we finally choose a list of strategies to compare with each other: $RD$, $AR$, $HWR$, $UCT_{100, 1}$, $UCT_{200, 2^{-5}}$, $UCT_{400, 2^{-2}}$, $UCT_{800, 2^{-1}}$, and $UCT_{1600, 2^{-1}}$. The simulation results are summarized in Table~\ref{tab:mcts}, in which each cell contains the mean predicted win rate of the strategy displayed on the respective row. 

We define the strength of a strategy as the number of strategies it can defeat with more than $50\%$ win rate. Therefore, the weakest to strongest strategies are: $RD$, $AR$, $UCT_{100, 1}$, $UCT_{200, 2^{-5}}$, $HWR$, $UCT_{400, 2^{-2}}$, $UCT_{800, 2^{-1}}$, and $UCT_{1600, 2^{-1}}$. Given a sufficient number of iterations ($\geq400$), UCT strategies can outperform all non-UCT strategies, which proves the effectiveness of our proposed algorithm. There is a general trend that UCT improves the win rate as the number of iterations increases. Specifically, $UCT_{1600, 2^{-1}}$ can beat $HWR$ with a 61.1\% win rate, highest one among the other UCT-based strategies with fewer iterations. However, we can observe the phenomenon of diminishing gain as the number of iterations exceeds a certain threshold: $UCT_{1600, 2^{-1}}$ has a $51.3\%$ win rate against $UCT_{800, 2^{-1}}$, only marginally better than $50\%$ but with double the number of iterations.

Among non-UCT strategies, the $HWR$ strategy that always picks the highest win rate hero achieves the best performance, which can defeat UCT with 200 iterations with a $51.6\%$ win rate. However, $HWR$ cannot prevail over UCT strategies that are allowed with more than 200 iterations. $AR$ adopted from~\cite{hanke2017reco} defeats $RD$ with $68.2\%$ win rate for our implementation, while the original authors reported a $76.4\%$ win rate against $RD$. The discrepancy may be due to different datasets being used to mine association rules and train the win rate predictor.

To show the efficiency of different strategies, we report the average wall time (i.e. real elapsed time) needed per pick in Table~\ref{tab:mcts_time}. The time is recorded excluding the first pick, since it is based on weighted random sampling. UCT-based strategies take 1.12 seconds or less per pick, which is a small fraction of the 25 second time limit per pick for human players in real matches~\cite{dotapickorder}. This demonstrates the feasibility of applying the UCT-based hero recommendation system online in large scale and real time.

\begin{figure*}
\centering
\includegraphics[width=0.8\textwidth]{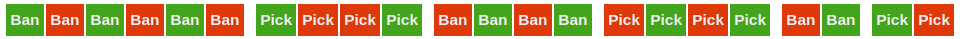}
\caption{Drafting order in match mode "Captain Mode" in DOTA 2. The green and red cells represent the turns of two different teams (i.e., Radiant and Dire), assuming the green team starts first. Diagram is adapted from~\cite{dotapickorder} and best viewed in color.}
\label{fig:pickorder_captain_mode}
\end{figure*}

\subsection{Extension to Other Match Modes}\label{sec:extension}
What we have focused on so far in this paper is based on the drafting rules of match mode \textit{All Ranked} in DOTA 2. However, our MCTS-based algorithm is not limited solely to this mode and can be adapted to other modes, whereby additional drafting rules and mechanics are deployed. In this section, we detail how our algorithm can be extended to another popular match mode called \textit{Captain Mode} to demonstrate our algorithm's generality. 

During the drafting process in Captain Mode matches, there is another type of action, called \textit{hero banning}, that players can take besides hero pick. In the turns of banning, a team can designate certain heroes to be prohibited from selection by either team. The drafting order is represented in Figure~\ref{fig:pickorder_captain_mode}: two teams first alternate to ban three heroes, then alternate to pick two heroes, then alternate to ban two heroes, and so on. As the result, the drafting is 22 steps long, instead of just 10 as in the case of \textit{All Ranked}. The interleaving nature between bans and picks adds another level of complexity to the drafting, as players need to consider more possible strategies to prevent opponents from selecting their desired heroes. 

Despite being more complex, the drafting in Captain Mode matches can be formulated as a combinatorial game similar to that in All Ranked matches but with a few minor adjustments:

\begin{itemize}[leftmargin=*]
    \item The components of a game state $s$ can take an additional value of a special symbol $\Xi$ denoting corresponding banned heroes, i.e.,:
\begin{equation}
s_{i}=
\begin{cases}
  1, & \text{hero } i \text{ picked by Radiant} \\
  -1, & \text{hero } i \text{ picked by Dire} \\
  \Xi, & \text{hero }i \text{ banned} \\
  0, & \text{otherwise}
\end{cases}
\label{eqn:sifeature}
\end{equation}

The terminal state set $S_T$ will also change accordingly. A terminal state $s \in S_T$ include five ones, five negative ones, and 12 special symbol $\Xi$.

    \item An action can be either a ban or a pick. If it is a ban, it changes a zero component of a non-terminal state $s \in S \setminus S_T$ to $\Xi$. 
    \item The turn function $\rho$ will also be updated to reflect the drafting order in Figure~\ref{fig:pickorder_captain_mode}.
    \item When a terminal state $s \in S_T$ is fed into the reward function $R(s)$, its components of $\Xi$ will be treated as zeroes.
\end{itemize}

Given the above formulation of the Captain Mode drafting, in the form of a combinatorial game, the UCT algorithm can be applied directly in the same way as described in Section~\ref{sec:mcts}. Within the drafting rules of Captain Mode matches, we conduct the same simulation-based experiments to compare UCT with the baselines $RD$, $AR$ and $HWR$. When taking the banning action, $AR$ is executed from the perspective of the opponent - identifying candidate heroes to ban as if the opponent is to pick; $HWR$ always chooses among the available heroes with the highest win rate for both the pick and ban actions. Regarding other implementation details, the selection frequency-weighted sampling is executed for the first hero to ban rather than the first hero to pick. $c$ becomes effective in $UCT_{100, c}$, since the simulation step may kick off after the starting root node is no longer expandable. We find the best $c$ to match with ${n=100}$ is $2^{-5}$, while the best $c$ values to couple with other ${n=\{200, 400, 800, 1600\}}$ remain the same as in All Ranked experiments.

The evaluation results are shown in Table~\ref{tab:mcts_captain_mode}. First, we observe that the strength order of the tested strategies remains the same, with $RD$ being the weakest and $UCT_{1600, 2^{-1}}$ being the strongest. Second, the win rate of UCT strategies against non-UCT strategies usually sees a 1-3\%  improvement in absolute value compared to the counterparts in All Ranked experiments. For example, $UCT_{1600, 2^{-1}}$ has a 62.7\% win rate against $HWR$ in Captain Mode, which is larger than the win rate $UCT_{1600, 2^{-1}}$ has against $HWR$ in All Ranked (61.1\%). This highlights the advantage of employing simulation-based approaches such as MCTS: that the advanced planning brought by the MCTS algorithm could further widen its gap over baseline methods in more sophisticated settings.

We also observed a negligible change in the average wall time needed per pick in Captain Mode-based simulations, as compared to All Ranked (Table~\ref{tab:mcts_time}), so we do not report it here. This implies that the overhead incurred by the deeper search in UCT is relatively smaller than that in other components, such as computing the reward function.

\section{Conclusions, Limitations and Future Works}

In this paper, we treat the drafting process in MOBA games as a combinatorial game. Under this view, we propose a recommendation system that can effectively and efficiently search for the hero pick optimal for team victory based on Monte Carlo Tree Search. We design and conduct simulation-based experiments based on two kinds of drafting rules, confirming that MCTS-based recommendations can lead to stronger hero line-ups in terms of predicted team win rates, as compared to other baselines.

One limitation of this work is that we do not consider player-specific information, such as player skills in their selected heroes, when recommending heroes. It is possible that a hero recommended by our algorithm, which is based solely on the current hero line-up, may be poorly played by a player who is not familiar with it. Our algorithm can be extended to integrate player skills, by augmenting the game state with player information and training a more advanced win rate predictor as the reward function which takes as input both hero picks and player-specific information. This limitation could be addressed when we have access to needed player-specific data.

There are some other promising future directions that we can pursue next. First, were hero pick sequences from real match data available, we would integrate them as prior information to improve the tree policy and default policy in MCTS, thereby improving capabilities to build search trees more effectively and efficiently~\cite{gelly2007combining,chaslot2009adding}. Second, we would also like to investigate how our recommendation systems can be customized to account for additional drafting rules and extended to other real-world scenarios such as player drafting in sports~\cite{staw1995sunk}. 



\bibliographystyle{ACM-Reference-Format}
\bibliography{sigproc,mcts,urls,mobas} 

\end{document}